\documentclass{article}
\usepackage{arxiv}

\usepackage[utf8]{inputenc} 
\usepackage[T1]{fontenc}    
\usepackage{url}            
\usepackage{booktabs}       
\usepackage{amsfonts}       
\usepackage{nicefrac}       
\usepackage{microtype}      
\usepackage{lipsum}
\usepackage{cellspace}
\usepackage{alltt}
\usepackage{pifont}
\usepackage{calc}
\usepackage{mathtools}
\usepackage{booktabs}
\usepackage{url}
\usepackage{caption}
\usepackage{longtable}
\usepackage{amsmath}
\usepackage{amsthm}
\usepackage{amssymb}
\usepackage{bm}
\usepackage{adjustbox}
\usepackage{wrapfig}
\usepackage{epstopdf}
\usepackage{amsmath}
\usepackage{tikz}
\usepackage{framed}
\usepackage{thmtools,thm-restate}
\usepackage{enumitem}
\usepackage{algorithmic}
\usepackage{algorithm}
\usepackage{soul}
\usepackage{tcolorbox}
\usepackage{multirow}
\usepackage{array}
\usepackage{caption}
\usepackage{subcaption}
\usepackage{makecell}
\usepackage[colorlinks=true,breaklinks]{hyperref}
\usepackage{dsfont}

\usepackage{pifont}

\usepackage{graphicx}

\usepackage{hyperref}
\hypersetup{
    colorlinks=true,
    linkcolor=blue,
    filecolor=magenta,      
    urlcolor=cyan,
    pdftitle={Overleaf Example},
    pdfpagemode=FullScreen,
}

\usepackage{cleveref}

\usetikzlibrary{arrows}
\usetikzlibrary{shapes,backgrounds}

\graphicspath{{./Images/}}

\newtheorem*{theorem*}{Theorem}

\newtheorem*{lemma*}{Lemma}

\makeatletter
\renewcommand{\th@definition}{%
  \normalfont
  \thm@preskip-2 \relax
  \thm@postskip-2 \relax
}
\makeatother

\title{Collaborative Multi-source Domain Adaptation Through Optimal Transport}

\author{Omar Ghannou$^{1,2}$, Youn{\`e}s Bennani$^{1}$\\
  $^{1}$Universit\'e Sorbonne Paris Nord\\
  LIPN, UMR 7030 CNRS \\
LaMSN - La Maison des Sciences Numériques\\
F-93210, Plaine Saint-Denis - France\\
  \url{name.surname@sorbonne-paris-nord.fr}\\
$^{2}$Aix-Marseille Université\\
LIS, UMR 7020 CNRS\\
F-13397, Marseille - France\\
  \url{omar.ghannou@univ-amu.fr}
  }

\begin{document}

\maketitle

\begin{abstract}
Multi-source Domain Adaptation (MDA) seeks to adapt models trained on data from multiple labeled source domains to perform effectively on an unlabeled target domain data, assuming access to sources data. To address the challenges of model adaptation and data privacy, we introduce Collaborative MDA Through Optimal Transport (CMDA-OT), a novel framework consisting of two key phases. In the first phase, each source domain is independently adapted to the target domain using optimal transport methods. In the second phase, a centralized collaborative learning architecture is employed, which aggregates the N models from the N sources without accessing their data, thereby safeguarding privacy. During this process, the server leverages a small set of pseudo-labeled samples from the target domain, known as the target validation subset, to refine and guide the adaptation. This dual-phase approach not only improves model performance on the target domain but also addresses vital privacy challenges inherent in domain adaptation.

\end{abstract}



\section{Introduction}

Supervised learning is a cornerstone of machine learning, but it relies heavily on labeled data, which is often expensive and time-consuming to obtain. While transfer learning and generalization offer potential solutions, domain shift—defined as the divergence between the probability distributions of two domains—remains a significant challenge. Consequently, a model trained on labeled data from domain A (source) and tested on unlabeled data from domain B (target) typically suffers from performance degradation compared to being trained and tested within the same domain B, which is impractical without labels. To address these challenges, domain adaptation methods are employed. Within these approaches, some can manage adaptation between a single source and a single target, referred to as single-source domain adaptation (SDA) methods, while others handle adaptation between multiple sources and a single target, known as multi-source domain adaptation (MDA) methods.

Several works have advanced MDA. For instance, \cite{Mansour-theory} provided a theoretical analysis based on the assumption that the target is a convex combination of source domains. \cite{Zhao-bounds} proposed new error bounds for regression and classification tasks. Other research efforts focused on specific algorithms, such as combining diverse classifiers through novel weighting techniques \cite{Zhao-combining}, or using domain transform frameworks to find latent domain spaces via clustering \cite{Hoffman}. Generally, these studies estimate a mixed distribution or combine multiple single-source models.

The collaborative learning paradigm generally, and federated learning particularly, has garnered significant attention from both academia and industry. Initially introduced by \cite{FL}, this approach enables model training on data from various sources, such as mobile devices or organizations, without requiring centralized data access. This concept is pivotal for addressing data privacy concerns in domain adaptation, as it allows for the utilization of source domain data without direct access.

Unsupervised domain adaptation methods based on optimal transport have recently gained traction due to their success in various machine learning problems \cite{laclau2017co} \cite{bouazza2019multi} \cite{Ben-Bouazza2022}. These methods are adept at discovering and learning the minimal transformation cost from source to target domains, effectively reducing domain discrepancies (i.e., domain shift).

In this context, we propose a novel approach that combines optimal transport with collaborative learning to enhance model performance on the target domain using multiple source domains. Our method, which we term Collaborative Multi-source Domain Adaptation through Optimal Transport (CMDA-OT), leverages the strengths of both frameworks to achieve superior adaptation while preserving data privacy.

The rest of this paper is structured as follows: Section 2 provides the necessary preliminary knowledge and notations for unsupervised multi-source domain adaptation and optimal transport theory. Section 3 presents our proposed approach for MDA, CMDA-OT. Section 4 demonstrates a comparative study with state-of-the-art methods on two benchmark datasets for MDA. Finally, Section 5 concludes the paper.

\vspace{-1mm}

\section{Fundamental background of the proposed approach}

\subsection{Unsupervised Multi-Source Domain Adaptation}

Let $\mathcal{X} = \mathbb{R}^d$ be an input space and $\mathcal{Y}=\{c_1,...,c_k\}$ a set of label space consisting of $k$ classes. Let's consider $\mathcal{S}_i \,  \forall i \in \{1,...,N\}$, $N$ distinct probability distributions over the product space $\mathcal{X} \times \mathcal{Y}$ called the source domains, and $\mathcal{T}$ a probability distribution over the product space $\mathcal{X} \times \mathcal{Y}$ called the target domain. 

Unsupervised Multi-source Domain Adaptation (UMDA) is a transductive learning problem. Both labeled data from the source domains and unlabeled data from the target domain are assumed to be accessible throughout the training phase \cite{redko2019advances}. 

More formally, for each source, we have access to a set $S_i=\{(x_j^i,y_j^i)\}_{j=1}^{n_i}$ of $n_i$ independent and identically distributed (i.i.d.) labeled samples of the $i^{th}$ source pulled from the joint distribution $\mathcal{S}_i$, and a set $T=\{x_j\}_{j=1}^m$ of $m$ i.i.d. unlabeled target samples pulled from the marginal distribution $\mathcal{T}_\mathcal{X}$ of the joint distribution $\mathcal{T}$ over $\mathcal{X}$:

\begin{center}
    $S_i=\{(x_j^i,y_j^i)\}^{n_i}_{j=1} \sim (\mathcal{S}_i)^{n_i}, \, \forall i \in \{1,...,N\}$ \quad \text{and} \quad $T=\{x_j\}_{j=1}^m \sim (\mathcal{T}_\mathcal{X})^m$.
\end{center}

The objective of unsupervised multi-source domain adaptation is to build a robust classifier $\eta: \mathcal{X} \to \mathcal{Y}$ with a low target risk:
\begin{center}
$\mathcal{R}_{\mathcal{T}}(\eta) = \underset{(x,y) \sim \mathcal{T}}{\mathbb{P}} (\eta(x) \neq y)$,
\end{center}
under the assumption of domain shift not only between the source and target domains but also among the source domains themselves:

\begin{center}
$\forall i,j \in \{1, .... , N\}, \, i \neq j \implies 
\mathcal{S}_{i} \neq \mathcal{S}_{j}$
\quad \text{and} \quad
$\forall i \in \{1, .... , N\}, \,
\mathcal{S}_i \neq \mathcal{T}$.
\end{center}

\subsection{Optimal Transport}\label{AA}

Let's consider two probability measures $\mu$ and $\nu$ over spaces $\mathcal{X}$ and $\mathcal{Y}$ respectively, and a positive cost function $c : \mathcal{X}\times\mathcal{Y} \to \mathbb{R}^+ $.
The aim of optimal transport is to minimize the total cost of transporting the measure
$\mu$ into $\nu$. 

In the discrete formulation of optimal transport, when $\mu$ and $\nu$ are only available through finite samples $X=\{x_1,x_2,..., x_n\}$ and $Y=\{y_1,y_2,..., y_m\}$, they can be considered as empirical distributions: 

\begin{equation} 
\mu = \sum_{i=1}^n \alpha_{i} \delta_{x_{i}} \,\, \text{and} \,\, \nu = \sum_{j=1}^m \beta_{j} \delta_{y_{j}} \label{eq6}
\end{equation}
where $\alpha=(\alpha_1,...,\alpha_n)$ and $\beta=(\beta_1,...,\beta_m)$ are vectors in the probability simplex $\sum_n$ and $\sum_m$ respectively. In this case, the cost function only needs to be specified for every pair $(x_i,y_j)_{i,j}$, so it can be compactly represented by a cost matrix $C \in  \mathcal{M}_{n \times m}(\mathbb{R}^{+})$. 

The formulation of Monge aims to get a measurable transport map $\texttt{T} : \mathcal{X} \to \mathcal{Y}$ that assigns each point $x_i$ to a single point $y_j$, and which pushes the mass of $\mu$ toward that of $\nu$ (i.e., $\texttt{T}\#\mu = \nu$), while minimizing the total cost of transportation:

\begin{equation} (\mathcal{M}) \,\,\,\underset{\texttt{T}}{\min}\{\sum_{i=1}^n  c(x_i,\texttt{T}(x_i)) \,\, | \,\, \texttt{T}\#\mu = \nu \},  \label{eq7} \end{equation}
where $\#$ denotes the pushforward operator.

Since the problem of Monge may not admit a solution, a convex relaxation was suggested by Kantorovich that considers instead of the transport map $\texttt{T}$ a soft multi-valued assignment defined in terms of probabilistic couplings $\gamma \in \mathcal{M}_{n \times m}(\mathbb{R}^{+})$ whose marginals recover $\mu$ and $\nu$. Formally, Kantorovich’s formulation seeks $\gamma$ in the transportation polytope:

\begin{equation} 
U(\alpha,\beta) = \{\gamma \in \mathcal{M}_{n \times m}(\mathbb{R}^{+}) \, | \, \gamma \mathds{1}_{m} = \alpha \, \text{and} \, \gamma^{\mathrm{T}} \mathds{1}_{n} = \beta\} \label{eq8}
\end{equation}
that solves:
\begin{equation}
(\mathcal{MK})\,\,\,\underset{\gamma \in U(\alpha,\beta)}{\min}  \langle {\gamma},{C} \rangle _F, \label{eq9}
\end{equation}
where $\langle.,.\rangle_F$ is Frobenius inner product of matrices.

The discrete Kantorovich problem is a linear program and can therefore be solved exactly in $\mathcal{O}(r^3 \log(r))$, where $r= \max(n,m)$ using the simplex algorithm and interior point methods, which is computationally intensive. In practice, an entropy smoothing variant \cite{cuturi2013sinkhorn} leads to much more efficient algorithms:

\begin{equation}
(\mathcal{MK}^{\,\varepsilon}) \,\,\,\,\,\,\underset{\gamma \in U(\alpha,\beta)}{\min}  \langle {\gamma},{C} \rangle _F - \varepsilon \mathcal{H}(\gamma)   , \label{eq10}
\end{equation}
where $\mathcal{H}(\gamma) = - \sum_{i=1}^n \sum_{j=1}^m \gamma_{ij} (\log(\gamma_{ij}) - 1)  $ is the entropy of $\gamma$. The solution to this strictly convex optimization problem is of the form $\gamma^* = \text{diag}(u)K\text{diag}(v)$, with $K =\exp(\frac{-C}{\varepsilon})$, and can be obtained efficiently via the Sinkhorn-Knopp algorithm, an iterative matrix scaling method \cite{peyre2019computational}. 

\subsection{Collaborative Learning with Federated Learning as Instance}

\begin{figure}[!h]
\centerline{\includegraphics[width=0.8\linewidth]{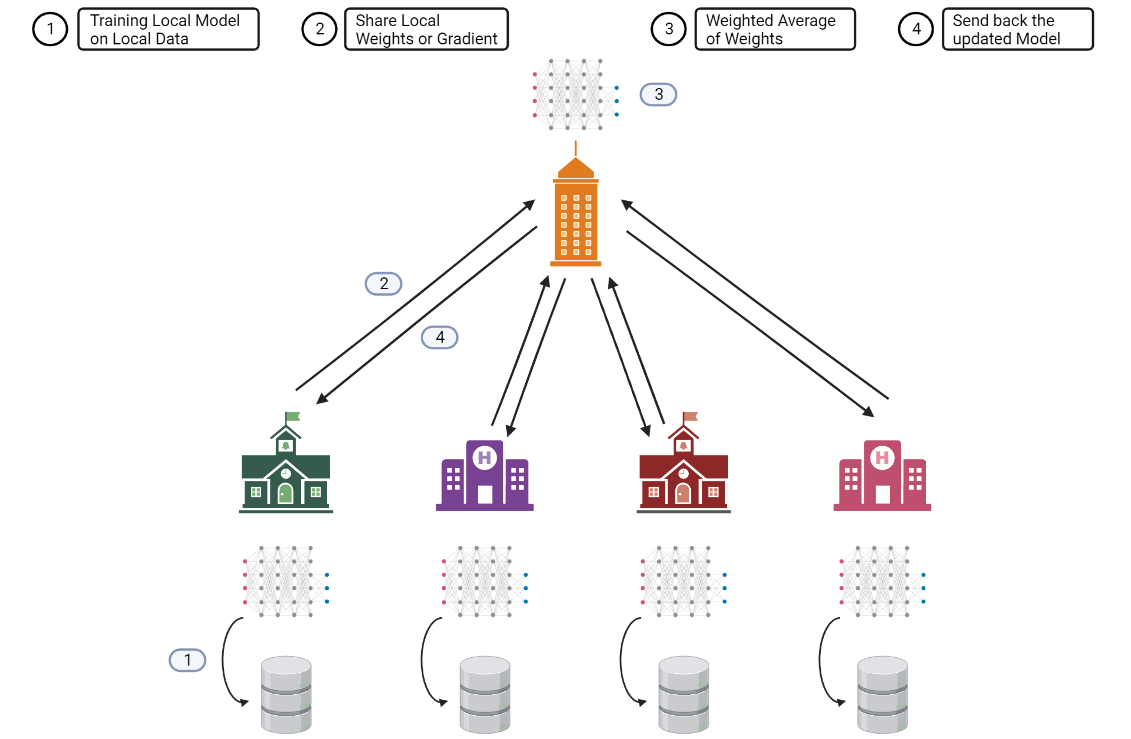}}
\caption{Centralized Federated Learning}
\label{centerFed}
\end{figure}

Collaborative learning enables the joint training of machine learning models through the collaboration of multiple parties (e.g., devices, organizations) without sharing local data. Federated learning (FL) is a specific instance of collaborative learning and is typically classified into either centralized or decentralized paradigms. Many contemporary FL systems support neural network training. Google proposed a scalable production system that allows the training of deep neural networks by hundreds of millions of devices \cite{FL-scale}. Centralized FL can be represented by a set of \(N\) models for \(N\) clients (i.e., data sources) and a server model that centralizes the learning process, as depicted in Figure \ref{centerFed}.

The training process in centralized FL consists of four repetitive phases. Firstly, each client model is trained on its local private data. Secondly, each client sends its model (i.e., weights) to the server, following the FedAvg algorithm \cite{FL}. Various algorithms, such as the FedSGD algorithm \cite{FL}, have been proposed to handle the exchange of information between models. The FedSGD algorithm allows for the transmission of gradients rather than weights, which is considered more effective but requires increased communication, potentially straining the distributed FL system. In the third phase, the server averages the received weights from the \(N\) clients. This average is weighted by the number of samples in each client's data, normalized by the sum of the counts of all samples across all clients, as proposed by Google \cite{FL}. Finally, in the fourth phase, the clients receive the new weights and update their models.\\

\begin{figure}[!htb]
\centerline{\includegraphics[width=0.8\linewidth]{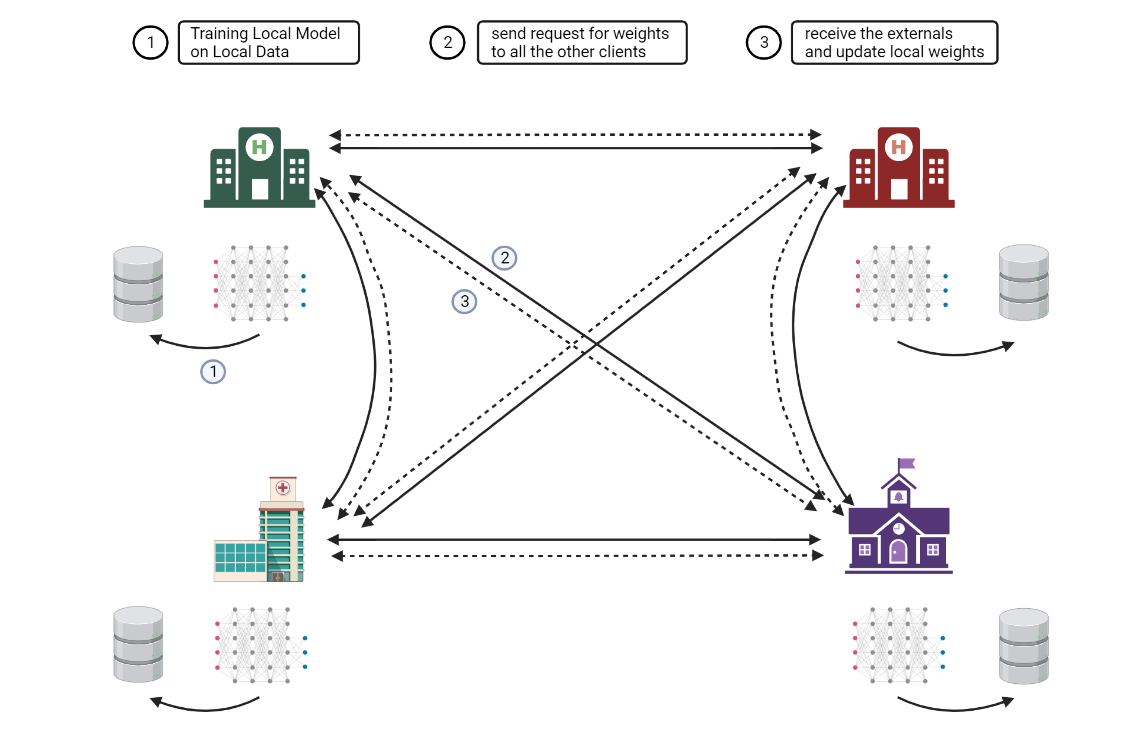}}
\caption{Decentralized Federated Learning}
\label{decenterFed}
\end{figure}

The decentralized or Peer-to-Peer federated learning paradigm follows the same principles as centralized FL, except it does not require a trusted server to centralize the learning process. Figure \ref{decenterFed} illustrates the overall repetitive process of training \(N\) client models.

Initially, each client trains its model on its local private data. In the second and third phases, it requests the weights or gradients from other clients to update its model with both its local data and the received exchangeables (i.e., weights or gradients). Numerous algorithms have been proposed within this paradigm. For instance, BrainTorrent \cite{BrainTorrent} is a selective algorithm where clients communicate and update their models using only the exchangeables from clients who have updates to send. This process continues until convergence. The order of operations is crucial in decentralized FL algorithms and significantly impacts the final training results. For example, the outcome may differ if client 1 uses the weights or gradients from client 2 before client 2 does. This challenge underscores the importance of the collaboration order. simFL \cite{simFL} is a decentralized approach to federated learning that adopts the centralized paradigm in a decentralized manner. This method involves selecting one client as the server in each iteration and treating the others as clients. The model is updated based on the chosen client's weights and the received ones before being sent back to the rest. However, simFL may encounter security issues or instability due to the variable server.


\section{Our Approach}

\subsection{Overall Framework of CMDA-OT}

\begin{figure}[!h]
\centering
\centerline{\includegraphics[width=0.98\linewidth]{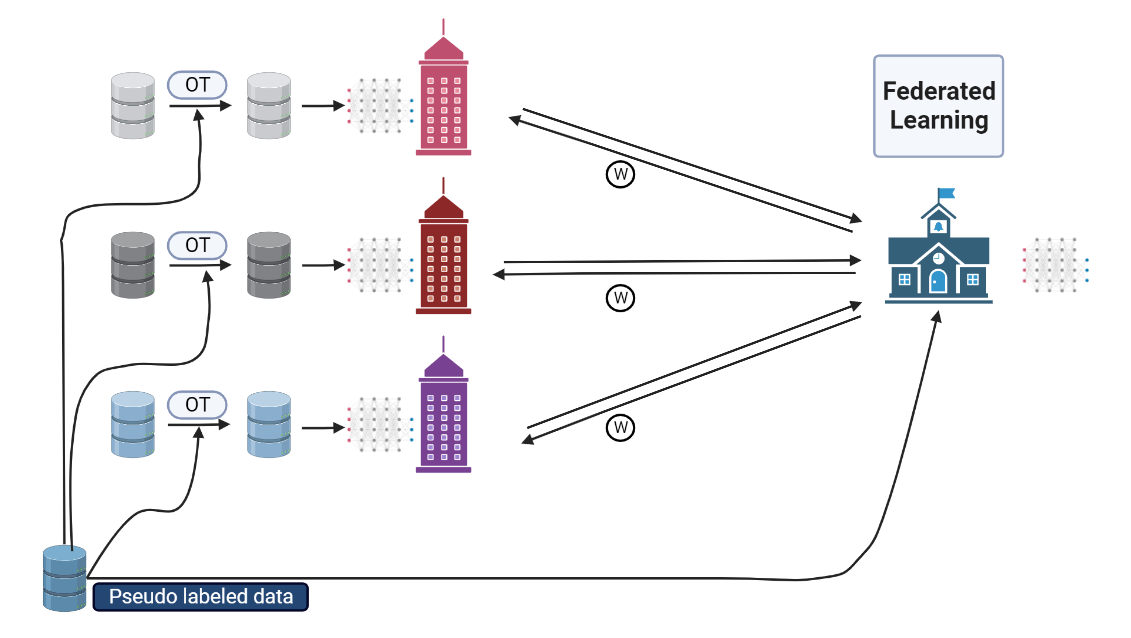}}
\caption{Overall Framework of CMDA-OT}
\label{overall}
\end{figure}

Figure \ref{overall} illustrates the overall collaborative framework of our proposition, which can be divided into two main parts of the adaptation. Initially, 
we apply optimal transport, as the first part, to project each source domain data into the target domain space using the Sinkhorn algorithm with L1L2 class regularization. This process aims to find a new representation of the source domain data that closely resembles the representation of the target domain, thereby reducing the domain shift phenomenon between them. This new representation is referred to as the transported data of the source domain.

Optimal transport generates new representations of the source domain data, which may not always be the most discriminative representation. In some cases, it could be inferior to the original representation of the source domain data, possibly due to factors such as data imbalance. To address this issue, we utilize a small set of pseudo-labeled data from the target domain, referred to as the target domain validation data, to assess the quality of the new representation. This involves testing and comparing the accuracy of models trained on both the transported and original data. If an improvement is observed when testing on the transported data, the client may opt to use this data for further steps of the adaptation. Otherwise, the original representation is retained as the preferred choice. 

The subsequent part involves the intervention of centralized collaborative learning, federated learning as instance, with each client acting as a source domain. In our approach, we employ the FedAvg algorithm \cite{FL} for the learning process. Here, each client sends its own weights after the local training step, and the server aggregates these weights, taking into account coefficients assigned to each client.

\cite{FL} defines the weight aggregation that takes place on the server as a weighted sum of all the clients' weights. The coefficient assigned to each client is directly proportional to the number of samples in that client. Formally:

\begin{equation}
W_s = \sum_{i=1}^{N} \frac{n_i}{S_N} W_i
\end{equation}

with:

\begin{equation}
S_N = \sum_{i=1}^{N} n_i
\end{equation}

While it is logical and sensible to weight the sum of the weights based on the number of samples provided by each client, this approach appears to work better in scenarios where all clients have the same probability distribution (i.e., the same domain). In Unsupervised Multi-source Domain Adaptation (UMDA), where different domains are involved, a client with fewer samples may contribute more effectively to improving the final model than another client with more data. Failing to account for this domain shift and weighting solely based on the number of samples can lead to less reliable results. To address this issue, we propose weighting all clients by defining coefficients that are directly proportional to the performance of each client model when tested on the small pseudo-labeled validation portion of target domain data. This test takes place in each client.

Our proposition takes into account the paramount importance of privacy, which is a necessity across all industries today. It is often challenging to train a model without direct access to the data, particularly in the context of domain adaptation and phenomena such as domain shift, where data from multiple sources may be required to leverage information for building a robust model while still adhering to privacy regulations. To enhance domain adaptation within the constraints of privacy, we propose CMDA-OT. Table \ref{tab1} highlights the advantages of CMDA-OT in comparison to relevant literature and related works.

\begin{table}[!htb]
\renewcommand{\arraystretch}{1.2}
\caption{Adaptation settings comparison}
\begin{center}
\begin{tabular}{|p{5.65cm}|p{2cm}|p{2cm}|p{2cm}|p{2cm}|}
\hline
\centering\textbf{}&\multicolumn{4}{|c|}{\textbf{Settings}} \\
\cline{2-5}
\centering\textbf{Method Name} & \textbf{\textit{Unlabelled Target Data}} & 
\centering\textbf{\textit{No Source Data Access}} & \centering\textbf{\textit{Source Model}} &
\textbf{\textit{Multiple Domains}}
\\
\hline
\centering HTL \cite{HTL}& 
\centering&
\centering\checkmark&
\centering\checkmark&
\\
\hline

\centering UDA \cite{UDA}& 
\centering\checkmark&
\centering&
\centering\checkmark&
\\
\hline

\centering MSDA \cite{MSDA}& 
\centering\checkmark&
\centering&
\centering\checkmark&
\checkmark\\
\hline

\centering U-HTL \cite{U-HTL}& 
\centering\checkmark&
\centering\checkmark&
\centering\checkmark&
\\
\hline

\centering STEM \cite{STEM}& 
\centering \checkmark&
\centering&
\centering \checkmark&
\checkmark\\
\hline

\centering MOST \cite{MOST}& 
\centering \checkmark&
\centering&
\centering \checkmark&
\checkmark\\
\hline

\centering CMDA-OT (Ours)& 
\centering \checkmark&
\centering \checkmark&
\centering \checkmark&
\checkmark\\
\hline
\end{tabular}
\label{tab1}
\end{center}
\end{table}

\subsection{Pseudo-Labeling the Target Domain Validation Data}

\begin{figure}[!h]
\centerline{\includegraphics[width=0.4\linewidth]{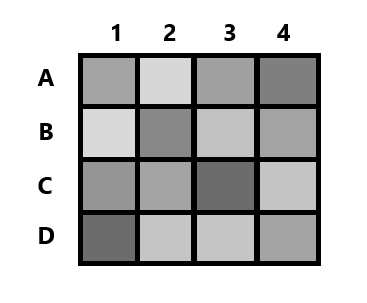}}
\caption{HOT correspondence Matrix}
\label{HOT}
\end{figure}

The significance of this validation proportion lies in its ability to guide adaptation, thereby reducing domain shift by maintaining a dynamic training process. In our scalable proposition, this validation proportion can be labeled by experts if available, which can lead to improved results. Alternatively, we propose a pseudo-labeling approach to handle fully unsupervised training for the target domain model. The HOT \cite{HOT} approach offers a scalable solution that can be applied in our scenario. Initially, we cluster the target domain data on the server using one of the clustering algorithms proposed by the literature. In our experiments, we utilized spectral clustering, as it represents one of the state-of-the-art in clustering algorithms. Once the clusters are obtained and random labels are assigned, this pseudo-labeled validation data is distributed to each client. Here, we apply hierarchical optimal transport (HOT), which calculates the correspondence between each source domain labels and the random labels assigned to this data during clustering. Figure \ref{HOT} illustrates an example of the HOT correspondence matrix.

The results obtained from different clients are not always identical. Therefore, when there are three or more clients, we apply a majority vote to determine the outcome. In cases where there are only two clients, we use the correspondence of the source domain that is closest to the target domain, as determined by the Wasserstein distance.

\subsection{On-Client Testing and Server Weights Aggregation}

The validation part of the target data serves as a guide for adaptation in both phases. During the optimal transport phase, this part determines whether accuracy improves when using the new representation obtained by the optimal transport transformation. If not, the original data representation is retained. This testing occurs across all clients, i.e., all sources. Each client receives this pseudo-labeled data and conducts an internal test to determine which representation, i.e., original or transported, to use in the next phase of adaptation, which is the collaborative learning.

During the collaborative learning training phase, the server utilizes the target domain validation portion. Upon receiving the models weights from all clients, the server tests each client's model on this validation set and records their accuracies. This process is repeated for all client models. The accuracies are then normalized by their sum and used as coefficients for aggregating the incoming models weights. This approach ensures a highly dynamic adaptation, continuously adjusting the target model as each client's model evolves during the learning process. It also provides a robust metric for determining the importance of each source to the target domain.

The effectiveness of validation data pseudo-labeling is directly linked to the quality of clustering. Higher cluster purity results in more accurate label assignments. Therefore, we employed spectral clustering in our experiments. The HOT \cite{HOT} method effectively maps random labels from clustering to real labels. This approach ensures that the training process remains entirely unsupervised for better fine-tuning the target model weights, while being flexible enough to replace the expert-labeled validation set. When clusters are pure, the pseudo labels closely match the actual labels.

In addition to preserving privacy, collaborative learning provides substantial benefits in scalability and flexibility. By sharing the weights of each source model, we overcome the limitations of individual models when aggregating them into a unified target model. This collaborative approach allows us to fully leverage the potential of data and knowledge without sharing or exposing the actual data or compromising privacy.

\subsection{Collaborative Learning Framework : Adaptability and Scalability}

This framework leverages the principles of optimal transport and collaborative learning, and although we have implemented it using centralized federated learning as an instance, the framework is designed to be adaptable to various collaborative learning paradigms. In the centralized federated learning paradigm, our approach ensures that we do not have direct access to the client data (i.e., source domain data), thus preserving client privacy. Instead, the federated server manages the learning process and has access to the target data. This implementation demonstrates the versatility of our framework, allowing it to be extended to other collaborative learning settings while maintaining robust privacy-preserving mechanisms.
\vspace{-2mm}
\section{Experiments}
\subsection{Datasets}

We conducted our experiments using two well-established benchmark datasets: Office-Caltech-10 and VLSC.

\textbf{VLSC:} This dataset includes images from four different domains: Caltech101, VOC2007, SUN09, and LabelMe. Each domain contains five classes: birds, dogs, chairs, cars, and humans.

\textbf{Office-Caltech10:} This collection consists of images from four domains: Amazon, DSLR, Webcam, and Caltech. Each domain comprises 10 categories commonly encountered in daily life, household, and office settings, such as computers, bags, cars, and bikes.

\subsection{Hyperparameters Tuning}
In our experiments, the hyperparameters were selected using a test-driven approach to ensure the performance. We also provide recommendations for tuning these parameters in the following discussion, as they can vary depending on the data distribution. Specifically, during our experiments, we set the entropy regularization to 50 and the class regularization to 5000 for the optimal transport method. For the clustering algorithm (spectral clustering), the number of neighbors was set to 12. It's important to note that these parameters are inherently sensitive to the underlying data distribution, and adjustments may be necessary depending on the specific characteristics of the data used.

\subsection{Results and Discussion}

In all experiments, we evaluated each model on each target multiple times, and relied on the average performance. Currently, the number of tests conducted is 1000.

\begin{table}[!h]
\renewcommand{\arraystretch}{1.2}
\caption{CMDA-OT results on VLSC}
\begin{center}
\begin{tabular}{|p{2.8cm}|p{0.8cm}|p{0.8cm} |p{0.8cm}|p{0.8cm}|p{0.8cm}|}
\hline
\centering\textbf{}&\multicolumn{4}{|c|}{\textbf{Target}}&\multicolumn{1}{|c|}{\textbf{}}
\cr\cline{2-5}
\centering\textbf{Method Name} & \centering$\rightarrow$\textbf{\textit{V}} &  \centering$\rightarrow$\textbf{\textit{L}} & \centering$\rightarrow$\textbf{\textit{S}} & \centering$\rightarrow$\textbf{\textit{C}} &
\textbf{AVG}
\\
\hline
\centering CMSD \cite{CMSD}& 
\centering 37.30& 
\centering 52.74& 
\centering 34.96& 
\centering 31.02& 
 39.00\\
\hline

\centering DS \cite{CMSD}& 
\centering 39.52& 
\centering 51.44& 
\centering 41.87& 
\centering 36.44& 
 42.31\\
\hline

\centering TCA+CMSD \cite{CMSD}& 
\centering 65.08& 
\centering 54.63& 
\centering 56.69& 
\centering 80.84& 
 64.31\\
\hline

\centering TCA+WAF \cite{CMSD}& 
\centering 66.67& 
\centering 54.81& 
\centering 56.40& 
\centering 80.53& 
 64.60\\
\hline

\centering TCA+WDS \cite{CMSD}& 
\centering 65.68& 
\centering 56.71& 
\centering 59.71& 
\centering 81.22& 
 65.83\\
\hline

\centering TCA+WDSC \cite{CMSD}& 
\centering 65.82& 
\centering 59.69& 
\centering 61.76& 
\centering 81.00& 
 67.05\\
\hline

\centering CMDA-OT (Ours)& 
\centering \textbf{69.0}& 
\centering \textbf{61.0}& 
\centering \textbf{67.0}& 
\centering \textbf{96.0}& 
\textbf{73.25}\\
\hline

\end{tabular}
\label{tab2}
\end{center}
\end{table}

\begin{table}[!htb]
\renewcommand{\arraystretch}{1.2}
\caption{CMDA-OT results on Caltech-Office-10}
\begin{center}
\begin{tabular}{|p{2.8cm}|p{0.8cm}|p{0.8cm} |p{0.8cm}|p{0.8cm}|p{0.8cm}|}
\hline
\centering\textbf{}&\multicolumn{4}{|c|}{\textbf{Target}}&\multicolumn{1}{|c|}{\textbf{}}
\cr\cline{2-5}
\centering\textbf{Method Name } & \centering$\rightarrow$\textbf{\textit{C}} &  \centering$\rightarrow$\textbf{\textit{A}} & \centering$\rightarrow$\textbf{\textit{W}} & \centering$\rightarrow$\textbf{\textit{D}} &
\textbf{AVG}
\\
\hline
\centering ResNet-101 \cite{RN}& 
\centering 85.4& 
\centering 88.7& 
\centering 99.1& 
\centering 98.2& 
 92.9\\ 
\hline

\centering DAN \cite{DAN}& 
\centering 89.2& 
\centering 91.6& 
\centering 99.5& 
\centering 99.1& 
 94.8\\ 
\hline

\centering DCTN \cite{DCTN}& 
\centering 90.2& 
\centering 92.7& 
\centering 99.4& 
\centering 99.0& 
95.3\\
\hline

\centering MCD \cite{MCD}& 
\centering 91.5& 
\centering 92.1& 
\centering 99.5& 
\centering 99.1& 
95.6\\
\hline

\centering $M^{3}SDA$ \cite{MSDA} & 
\centering \textbf{92.2}& 
\centering 94.5& 
\centering \textbf{99.5}& 
\centering \textbf{99.2}& 
96.4\\
\hline

\centering CMDA-OT (Ours)& 
\centering 92.0& 
\centering \textbf{95.1}& 
\centering 99.4& 
\centering \textbf{99.2}& 
\textbf{96.5}\\
\hline

\end{tabular}
\label{tab3}
\end{center}
\end{table}

The system includes some parameters that need to be tuned for optimal performance. One key parameter is the number of neighbors in spectral clustering, which needs adjustment to enhance clustering quality. Metrics like the Silhouette index or other internal indices can be used for this evaluation. The regularization parameters in optimal transport, whether Entropy-Regularization or Class-Regularization, must also be adjusted for each source (preferable) or uniformly across all sources. Certain regularization values may lead to mathematical issues, such as division by zero. For our experiments, we provide recommendations regarding the regularization values for each source to enhance the new representation obtained by optimal transport. These parameters can be adjusted locally in each source using pseudo-labeled validation data (test-driven approach). Generally, it is preferable to minimize Entropy-Regularization and maximize Class-Regularization without encountering mathematical errors. Minimizing Entropy-Regularization below a certain threshold or maximizing the gap between Entropy-Regularization and Class-Regularization may result in mathematical operational errors. These mathematical errors are inherent to the optimal transport algorithm and beyond our control.

To assess the performance of our approaches, we use the Friedman test and Nemenyi test as recommended in \cite{edman1937use} using the $Autorank$ package \cite{Herbold2020}. The Friedman test is conducted to test the null hypothesis that all approaches are equivalent in terms of accuracy. If the null hypothesis is rejected, the Nemenyi test is performed. If the average ranks of two approaches differ by at least the critical difference (CD), their performances are significantly different. In the Friedman test, we set the significance level at ${\alpha = 0.05}$.

\begin{figure}[!h]
\centerline{\includegraphics[width=\linewidth]{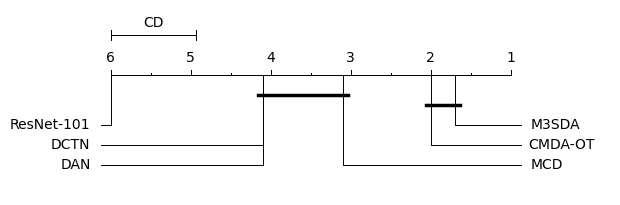}}
\caption{Friedman and Nemenyi test for comparing multiple approaches over Office-Caltech10
data sets: Approaches are ordered from right (the best) to left (the worst)}
\label{statOffc}
\end{figure}

\begin{figure}[!h]
\centerline{\includegraphics[width=\linewidth]{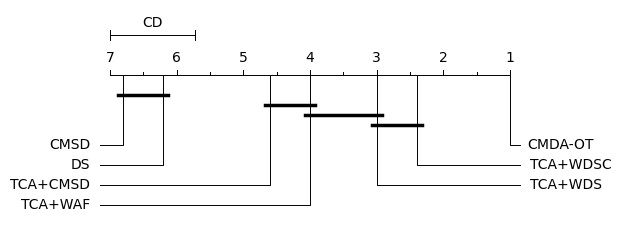}}
\caption{Friedman and Nemenyi test for comparing multiple approaches over VLSC
data sets: Approaches are ordered from right (the best) to left (the worst)}
\label{statVLSC}
\end{figure}

Figure \ref{statOffc} shows a critical diagram representing the projection of average ranks of approaches on an enumerated axis. The approaches are ordered from left (worst) to right (best), with a thick line connecting the approaches where the average ranks are not significantly different at a 5\% significance level. As shown in Figure \ref{statVLSC}, CMDA-OT achieves significant improvement over other proposed techniques, demonstrating stability during the federated learning phase. This process stops the collaboration for certain views when their local quality starts to decrease, preventing common issues in federated approaches. Compared to the most cited state-of-the-art approaches, the positive impact of using federated learning based on this theory is evident.

The statistical analysis was conducted for 7 approaches with 5 paired samples.
The family-wise significance level of the tests is alpha=0.050.
We rejected the null hypothesis that the population is normal for the approaches CMSD (p=0.000), DS (p=0.000), TCA+CMSD (p=0.000), TCA+WAF (p=0.000), TCA+WDS (p=0.000), TCA+WDSC (p=0.000), and CMDA-OT (p=0.000). Therefore, we assume that not all approaches are normal.
Because we have more than two approaches and some of them are not normal, we use the non-parametric Friedman test as omnibus test to determine if there are any significant differences between the median values of the approaches. We use the post-hoc Nemenyi test to infer which differences are significant.

\begin{table}[!h]
\centering
\resizebox{\columnwidth}{!}{
\begin{tabular}{lrrllll}
\toprule
 & MR & MED & MAD & CI & $\gamma$ & Magnitude \\
\midrule
ResNet-101 & 6.000 & 92.900 & 5.300 & [88.700, 98.200] & 0.000 & negligible \\
DAN & 4.100 & 94.800 & 4.300 & [91.600, 99.100] & -0.266 & small \\
DCTN & 4.100 & 95.300 & 3.700 & [92.700, 99.000] & -0.354 & small \\
MCD & 3.100 & 95.600 & 3.500 & [92.100, 99.100] & -0.405 & small \\
CMDA-OT & 2.000 & 96.500 & 2.700 & [95.100, 99.200] & -0.577 & medium \\
M3SDA & 1.700 & 96.400 & 2.800 & [94.500, 99.200] & -0.557 & medium \\
\bottomrule
\end{tabular}}
\caption{Summary of methods}
\label{tbl:stat_results}
\end{table}

\begin{table}[!h]
\centering
\resizebox{\columnwidth}{!}{
\begin{tabular}{lrrllll}
\toprule
 & MR & MED & MAD & CI & $\gamma$ & Magnitude \\
\midrule
CMSD & 6.800 & 37.300 & 2.340 & [34.960, 39.010] & 0.000 & negligible \\
DS & 6.200 & 41.870 & 2.350 & [39.520, 42.320] & -1.314 & large \\
TCA+CMSD & 4.600 & 64.310 & 7.620 & [56.690, 65.080] & -3.232 & large \\
TCA+WAF & 4.000 & 64.600 & 8.200 & [56.400, 66.670] & -3.054 & large \\
TCA+WDS & 3.000 & 65.680 & 5.970 & [59.710, 65.830] & -4.222 & large \\
TCA+WDSC & 2.400 & 65.820 & 4.060 & [61.760, 67.050] & -5.805 & large \\
CMDA-OT & 1.000 & 69.000 & 4.250 & [67.000, 73.250] & -6.233 & large \\
\bottomrule
\end{tabular}}
\caption{Summary of methods}
\label{tbl:stat_results}
\end{table}

We report the median (MD), the median absolute deviation (MAD) and the mean rank (MR) among all approaches over the samples. Differences between approaches are significant, if the difference of the mean rank is greater than the critical distance CD=1.274 of the Nemenyi test.
We reject the null hypothesis (p=0.000) of the Friedman test that there is no difference in the central tendency of the approaches CMSD (MD=37.300+-2.025, MAD=2.340, MR=6.800), DS (MD=41.870+-1.400, MAD=2.350, MR=6.200), TCA+CMSD (MD=64.310+-4.195, MAD=7.620, MR=4.600), TCA+WAF (MD=64.600+-5.135, MAD=8.200, MR=4.000), TCA+WDS (MD=65.680+-3.060, MAD=5.970, MR=3.000), TCA+WDSC (MD=65.820+-2.645, MAD=4.060, MR=2.400), and CMDA-OT (MD=69.000+-3.125, MAD=4.250, MR=1.000). Therefore, we assume that there is a statistically significant difference between the median values of the approaches.
Based on the post-hoc Nemenyi test, we assume that there are no significant differences within the following groups: CMSD and DS; TCA+CMSD and TCA+WAF; TCA+WAF and TCA+WDS; TCA+WDS and TCA+WDSC. All other differences are significant.

\begin{figure}[!h]
 \centering
 \begin{subfigure}{\textwidth}
 \centering
 \includegraphics[width=0.75\linewidth]{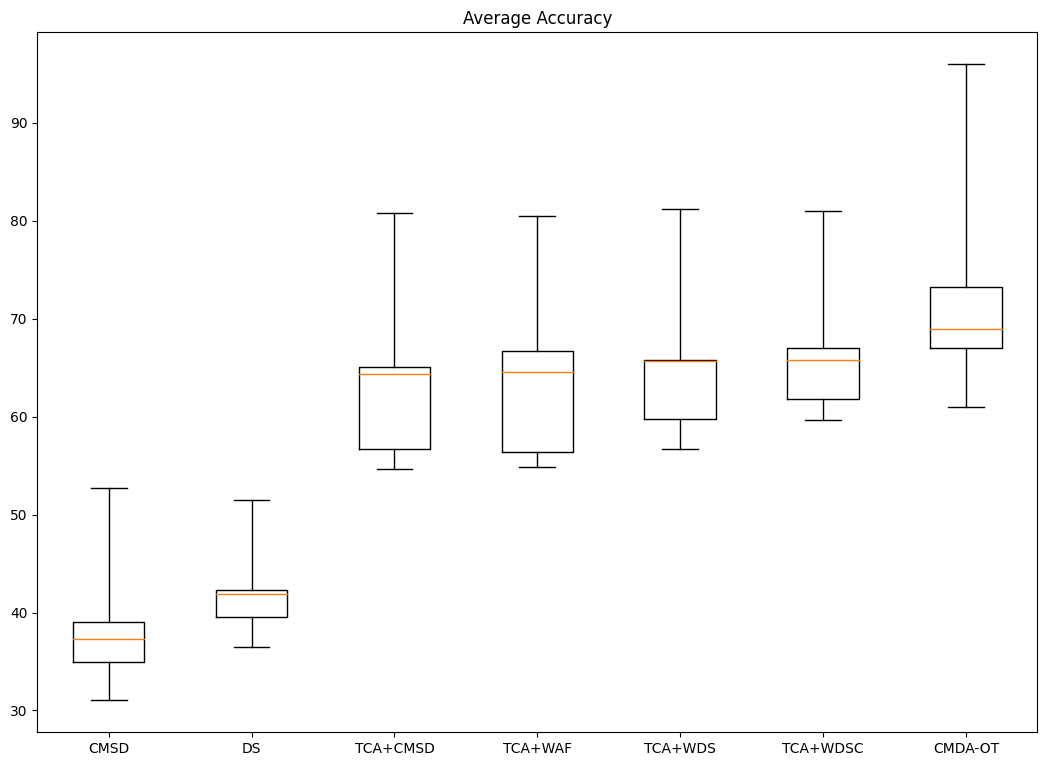} \caption{VLSC datasets}
 \end{subfigure}
 \begin{subfigure}{\textwidth}
 \centering
 \includegraphics[width=0.75\linewidth]{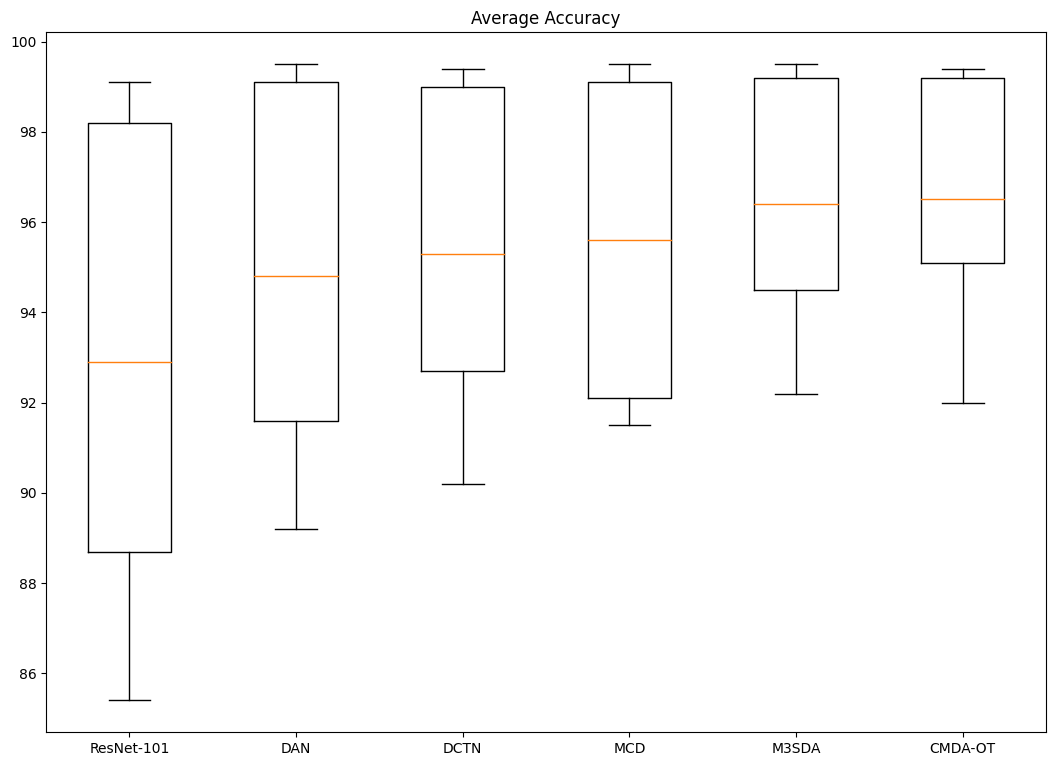} \caption{Office-Caltech10 datasets}
 \end{subfigure}

\caption{Sensitivity Box-Whiskers plots for the all approaches}
 \label{hboxes}
 \end{figure}

Sensitivity Box-Whiskers plots (figure \ref{hboxes}) represents a synthesis of the scores into five crucial pieces of information identifiable at a glance: position measurement, dispersion, asymmetry and length of Whiskers. The position measurement is characterised by the dividing line on the median (as well as the middle of the box). Dispersion is defined by the length of the Box-Whiskers (as well as the distance between the ends of the Whiskers and the gap). Asymmetry is defined as the deviation of the median line from the centre of the Box-Whiskers from the length of the box (as well as by the length of the upper Whiskers from the length of the lower Whiskers,  and by the number of scores on each side). The length of the Whiskers is the distance between the ends of the Whiskers in relation to the length of the Box-Whiskers (and the number of scores specifically marked).
These graphs show the same overall performance behaviour observed in the case of VLSC and Office-Caltech10 datasets. They show a clear improvement as a result of the the proposed approach. This improvement is observed for all datasets used.

\section*{Conclusion \& Future Direction}

In this work, we propose an architecture designed to ensure privacy preservation and facilitate domain adaptation in multi-source domain adaptation scenarios. Our approach enables unsupervised model training without access to the data sources, which is crucial for maintaining client privacy across diverse industries today. The architecture operates in two adaptation phases: initially, it uses selective optimal transport for each source, and subsequently, the collaborative learning, with Federated Learning (FL) server directs the adaptation by determining the significance of each source. Our method has proven effective on benchmark datasets. Looking ahead, there are several promising avenues for further enhancement and privacy preservation. Firstly, we believe that the performance of our architecture can be improved by developing or integrating more advanced collaborative or FL systems. Secondly, the new representations obtained via optimal transport at each source can be substituted with alternative representations from other algorithms that may offer superior discriminative capabilities.





\bibliographystyle{apalike}
\bibliography{references}

\end{document}